\def\year{2018}\relax
\documentclass[letterpaper]{article} %DO NOT CHANGE THIS
\usepackage{aaai18}  %Required
\usepackage{times}  %Required
\usepackage{helvet}  %Required
\usepackage{courier}  %Required
\usepackage{url}  %Required
\usepackage{graphicx}  %Required
\usepackage[dvipsnames]{xcolor}
\usepackage{tikz}
\usepackage{pgfplots}
\pgfplotsset{compat=1.9}
\usepackage{latexsym}
\usepackage{amsmath,amssymb}
\usepackage{rotating}
\usepackage{subcaption}
\usepackage{pdfpages}
\usepackage{booktabs}

\allowdisplaybreaks

\begin{document}
% The file aaai.sty is the style file for AAAI Press 
% proceedings, working notes, and technical reports.
%
%\title{Few-Shot Learning of Description Logic Concepts}
\title{Modeling Semantic Relatedness using Global Relation Vectors}
\author{Shoaib Jameel, Zied Bouraoui, Steven Schockaert}
\author{
Shoaib Jameel\\
Cardiff University, UK \\
JameelS1@Cardiff.ac.uk\\
\And
Zied Bouraoui\\
Cardiff University, UK\\
BouraouiZ@Cardiff.ac.uk\\
\And
Steven Schockaert\\
Cardiff University, UK\\
SchockaertS1@Cardiff.ac.uk
}

\maketitle
\begin{abstract}
Word embedding models such as GloVe rely on co-occurrence statistics from a large corpus to learn vector representations of word meaning. These vectors have proven to capture surprisingly fine-grained semantic and syntactic information. While we may similarly expect that co-occurrence statistics can be used to capture rich information about the relationships between different words, existing approaches for modeling such relationships have mostly relied on manipulating pre-trained word vectors. In this paper, we introduce a novel method which directly learns relation vectors from co-occurrence statistics. To this end, we first introduce a variant of GloVe, in which there is an explicit connection between word vectors and PMI weighted co-occurrence vectors. We then show how relation vectors can be naturally embedded into the resulting vector space.
\end{abstract}

\section{Introduction}

Word embeddings are vector space representations of word meaning \cite{DBLP:conf/nips/MikolovSCCD13,glove2014}. 
%In particular, most word embedding models represent words as vectors, such that similar words are represented as similar vectors (e.g.\ in terms of cosine similarity or Euclidean distance). 
One of the most remarkable properties of these models is that they can capture a wide array of syntactic and semantic relations, beyond mere similarity. For example, \cite{DBLP:conf/nips/MikolovSCCD13} found that analogy questions of the form ``$a$ is to $b$ what $c$ is to ?'' can often be answered by finding the word $d$ that maximizes $\cos(w_b-w_a+w_c,w_d)$, where we write $w_x$ for the vector representation of a word $x$. Along similar lines, \cite{Vylomova2016} considered the problem of predicting word pairs $(s,t)$ that are related in a given way, using only the fact that the word pairs $(s_1,t_1),...,(s_n,t_n)$ have this relationship. They found that training a linear SVM classifier on the vector differences $w_{t_1}-w_{s_1},...,w_{t_n}-w_{s_n}$ was often effective. %Other remarkable findings include the fact that numerical attributes, such a country's GDP, can be predicted from word embeddings using a linear regression model \cite{gupta2015distributional} and the fact that word embeddings can be decomposed in interpretable subspaces \cite{derracAIJ,DBLP:conf/acl/RotheS16}. On the other hand, several authors have also highlighted limitations of this vector difference model \cite{DBLP:conf/naacl/GladkovaDM16}, and of word embeddings as a source of semantic knowledge more generally \cite{DBLP:conf/acl/RubinsteinLSR15}.

Intuitively, the word vector $w_a$ represents $a$ in terms of its most salient features. For example, $w_{\textit{paris}}$ implicitly encodes that Paris is located in France and that it is a capital city, whereas $w_{\textit{france}}$ encodes e.g.\ that France is a country. Most of the salient features in which Paris and France differ are related to the fact that the former is a capital city and the latter is a country, which is essentially why the `capital of' relation can be modeled in terms of the vector difference $w_{\textit{france}} - w_{\textit{paris}}$. Many relationships, however, are not (completely) determined by the salient features of the considered words. For example, the fact that Germany won the FIFA World Cup in Brazil is unlikely to be captured by word embeddings. Instead, a common strategy is to learn to extract such kind of relationships from sentences that explicitly state them \cite{mintz2009distant,DBLP:conf/pkdd/RiedelYM10,DBLP:conf/acl/SantosXZ15}. 

While relation extraction has traditionally relied on surface patterns, dependency parses, and other linguistic features, word embeddings are  now increasingly being used for this purpose. A vector $r_{st}$ representing the considered relation between the words $s$ and $t$ is then learned from the embeddings of the words that occur between $s$ and $t$, in sentences that contain both. One surprisingly effective strategy to obtain such a relation vector is to simply average the embeddings of the words in between $s$ and $t$ \cite{DBLP:conf/emnlp/WestonBYU13,DBLP:conf/ranlp/FanCHG15,DBLP:conf/conll/HashimotoSMT15}. A standard classifier can then be trained on the resulting vectors. Despite its conceptual simplicity, this method can outperform convolutional networks and related neural network based architectures \cite{DBLP:conf/conll/HashimotoSMT15}. 

In this paper, we propose a new method to learn such relation vectors. Inspired by the GloVe word embedding model, we derive these vectors using global co-occurrence statistics, essentially modeling how often each word $j$ appears in sentences that contain both $s$ and $t$. In particular, we first discuss a variant of GloVe, in which word vectors can be directly interpreted as smoothed PMI-weighted bag-of-words representations. We then represent relationships between words as weighted bag-of-words representations, using generalizations of PMI to three arguments, and learn vectors that correspond to smoothed versions of these representations. 

% why better than averaging
% also captures implicit relationships: example (e.g. country is ally of)
% focus on semantic relations (between nouns)

%****************************************************
\section{Related Work}

%\todo{cite \url{https://arxiv.org/pdf/1704.05958.pdf}}

There is a long tradition in the field of Natural Language Processing (NLP) on relation extraction \cite{hearst1992automatic}. From a learning point of view, the most straightforward setting is where we are given labeled training sentences, with each label explicitly indicating what relationship is expressed in the sentence. This fully supervised setting has been the focus of several evaluation campaigns, including as part of ACE \cite{DBLP:conf/lrec/DoddingtonMPRSW04} and at SemEval 2010  \cite{DBLP:conf/semeval/HendrickxKKNSPP10}. A key problem with this setting, however, is that labeled training data is hard to obtain. A popular alternative is to use known instances of the relations of interest as a form of distant supervision \cite{mintz2009distant,DBLP:conf/pkdd/RiedelYM10}. 
%This means that relation extraction systems can be trained without requiring any human annotations, as long as we have a knowledge base at our disposal with sufficient instances of the relations of interest. Among others, distant supervision based extraction methods have been used to find relation instances that are missing from knowledge graphs such as Freebase and Yago.
Some authors have also considered unsupervised relation extraction methods \cite{shinyama2006preemptive,banko2007open}, in which case the aim is essentially to find clusters of patterns that express similar relationships, although these relationships may not correspond to the ones that are needed for the considered application. Finally, several systems have also used bootstrapping strategies \cite{brin1998extracting,agichtein2000snowball,carlson2010toward}, where a small set of instances are used to find extraction patterns, which are used to find more instances, which can in turn be used to find better extraction patterns, etc.

Traditionally, relation extraction systems have relied on a variety of linguistic features, such as lexical patterns, part-of-speech tags and dependency parses. More recently, several neural network architectures have been proposed for the relation extraction problem. These architectures rely on word embeddings to represent the words in the input sentence, and manipulate these word vectors to construct a relation vector. Some approaches simply represent the sentence (or the phrase connecting the entities whose relationship we want to determine) as a sequence of words, and use e.g.\ convolutional networks to aggregate the vectors of the words in this sequence \cite{zeng2014relation,DBLP:conf/acl/SantosXZ15}. Another possibility, explored in \cite{socher2012semantic}, is to use parse trees to capture the structure of the sentence, and to use recursive neural networks (RNNs) to aggregate the word vectors in a way which respects this structure. A similar approach is taken in \cite{xu2015classifying}, where LSTMs are applied to the shortest path between the two target words in a dependency parse. A straightforward baseline method is to simply take the average of the word vectors \cite{DBLP:journals/cogsci/MitchellL10}. While conceptually much simpler, variants of this approach have obtained state-of-the-art performance for relation classification \cite{DBLP:conf/conll/HashimotoSMT15} and a variety of tasks that require sentences to be represented as a vector \cite{DBLP:conf/naacl/HillCK16}.

Given the effectiveness of word vector averaging, in \cite{DBLP:conf/acl/KenterBR16} a model was proposed that explicitly tries to learn word vectors that generalize well when being averaged. Similarly, the model proposed in \cite{DBLP:conf/conll/HashimotoSMT15} aims to produce word vectors that perform well for the specific task of relation classification.  %The variant of GloVe that we propose below is also meant to improve how well relations can be modelled. However, instead of optimizing word vectors to be averaged, our proposed word embedding variant allows us to directly learn a relation vector from global co-occurrence statistics. 
%However, in contrast to the aforementioned approaches, we do not obtain relation vectors by aggregating word vectors, but instead learn such vectors directly. 
The ParagraphVector method from \cite{le2014distributed} is related to the aformentioned approaches, but it explicitly learns a vector representation for each paragraph along with the word embeddings. However, this method is computationally expensive, and often fails to outperform simpler approaches \cite{DBLP:conf/naacl/HillCK16}.

%****************************************************
\section{Word Vectors as Low-Rank PMI Vectors}\label{secWordEmbeddings}

Our approach to relation embedding is based on a variant of the GloVe word embedding model \cite{glove2014}. In this section,  we first briefly recall the GloVe model itself, after which discuss our proposed variant. A key advantage of this variant is that it allows us to directly interpret word vectors in terms of the Pointwise Mutual Information (PMI), which will be central to the way in which we learn relation vectors.

\subsection{Background}
The GloVe model \cite{glove2014} learns a vector $w_i$ for each word $i$ in the vocabulary, based on a matrix of co-occurrence counts, encoding how often two words appear within a given window. Let us write $x_{ij}$ for the number of times word $j$ appears in the context of word $i$ in some text corpus. More precisely, assume that there are $m$ sentences in the corpus, and let $\mathcal{P}^l_i \subseteq \{1,...,n_l\}$ be the set of positions from the $l^{\textit{th}}$ sentence where the word $i$ can be found (with $n_l$ the length of the sentence).
%$S_1,...,S_m$ be the sentences in the corpus, and let $w^l_1...w^l_{n_l}$ be the sequence of words in sentence $S_l$ (where each word is represented by an integer ID). 
Then $x_{ij}$ is defined as follows:
\begin{align*}
\sum_{l=1}^m \sum_{p\in \mathcal{P}^l_i}\sum_{q\in \mathcal{P}^l_j} \textit{weight}(p,q)
%x_{ij} = | \{(p,q,l) \,:\,& w^l_p=i, w^l_q=j, l\in \{1,...,m\},\\
%&p,q\in \{1,...,n_l\}, |p-l|\in \{1,...,W\}\} | 
\end{align*}
where $\textit{weight}(p,q)=\frac{1}{|p-q|}$ if $0< |p-q| \leq W$, and $\textit{weight}(p,q)=0$ otherwise, where the window size $W$ is usually set to 5 or 10. 

The GloVe model learns for each word $i$ two vectors $w_i$ and $\tilde{w_i}$ by optimizing the following objective:
\begin{align*}
%\sum_i\sum_{\substack{j\\x_{ij}\neq 0}} f(x_{ij}) (w_i \cdot \tilde{w_j} + b_{w_i} + \tilde{b}_{w_j} - \log x_{ij})^2    
\sum_i\sum_{j: x_{ij}\neq 0} f(x_{ij}) (w_i \cdot \tilde{w_j} + b_i + \tilde{b_j} - \log x_{ij})^2    
\end{align*}
where $f$ is a weighting function, aimed at reducing the impact of rare terms, and $b_i$ and $\tilde{b_j}$ are bias terms. The GloVe model is closely related to the notion of pointwise mutual information (PMI), which is defined for two words $i$ and $j$ as $\textit{PMI}(i,j) = \log \big(\frac{P(i,j)}{P(i)P(j)}\big)$, where $P(i,j)$ is the probability of seeing the words $i$ and $j$ if we randomly pick a word position from the corpus and a second word position within distance $W$ from the first position. The PMI between $i$ and $j$ is usually estimated as follows:
$$
\textit{PMI}_X(i,j) = \log \left(\frac{x_{ij}x_{**}}{x_{i*}x_{*j}}\right)
$$
where $x_{i*} = \sum_j x_{ij}$, $x_{*j}=\sum_i x_{ij}$ and $x_{**} = \sum_i\sum_j x_{ij}$. In particular, it is straightforward to see that after the reparameterization given by $b_i \mapsto b_i + \log x_{i*} - \log x_{**}$ and $b_j \mapsto b_j + \log x_{*j}$, the GloVe model is equivalent to
\begin{align}
%\sum_i\sum_{\substack{j\\x_{ij}\neq 0}} f(x_{ij}) (w_i \cdot \tilde{w_j} + b_{w_i} + \tilde{b}_{w_j} - \log x_{ij})^2    
\sum_i\sum_{j: x_{ij}\neq 0} f(x_{ij}) (w_i \cdot \tilde{w_j} + b_i + \tilde{b_j} - \textit{PMI}_X(i,j))^2    \label{eqGloVeAsPMI}
\end{align}
%******************************************************************************************
\subsection{A Variant of GloVe}\label{secVariant}
In this paper, we will use the following variant of the formulation in \eqref{eqGloVeAsPMI}:
\begin{align}\label{eqGloVeOurVariant}
%\sum_i\sum_{\substack{j\\x_{ij}\neq 0}} f(x_{ij}) (w_i \cdot \tilde{w_j} + b_{w_i} + \tilde{b}_{w_j} - \log x_{ij})^2    
\sum_i\sum_{j\in J_i} \frac{1}{\sigma_j^2} (w_i \cdot \tilde{w_j} + \tilde{b_j} - \textit{PMI}_S(i,j))^2    
\end{align}
Despite its similarity, this formulation differs from the GloVe model in a number of important ways. First, we use smoothed frequency counts instead of the observed frequency counts $x_{ij}$. In particular, the PMI between words $i$ and $j$ is given as:
%$$
%\textit{PMI}_S(i,j) = \log \left(\frac{(x_{ij} + \alpha)(x_{**}+n^2 \alpha)}{(x_{i*}+n\alpha)(x_{*j}+n\alpha)}\right)
%$$
$$
\textit{PMI}_S(i,j) = \log \left(\frac{P(i,j)}{P(i)P(j)}\right)
$$
where the probabilities are estimated as follows:
\begin{align*}
P(i) &= \frac{x_{i*} + \alpha}{x_{**} +n \alpha}&
P(j) &= \frac{x_{*j} + \alpha}{x_{**} +n \alpha}\\
P(i,j) &= \frac{x_{ij} + \alpha}{x_{**} +n^2 \alpha}
\end{align*}
where $\alpha\geq 0$ is a parameter controlling the amount of smoothing and $n$ is the size of the vocabulary. This ensures that the estimation of $\textit{PMI}(i,j)$ is well-defined even in cases where $x_{ij}=0$, meaning that we no longer have to restrict the inner summation to those $j$ for which $x_{ij}>0$. For efficiency reasons, in practice, we only consider a small subset of all context words $j$ for which $x_{ij}=0$, which is similar in spirit to the use of negative sampling in Skip-gram \cite{DBLP:conf/nips/MikolovSCCD13}. In particular, the set $J_i$ contains each $j$ such that $x_{ij}>0$ as well as $M$ uniformly\footnote{While the negative sampling method used in Skip-gram favors more frequent words, initial experiments suggested that deviating from a uniform distribution almost had no impact in our setting.} sampled context words $j$ for which $x_{ij}=0$, where we choose $M = 2 \cdot |\{j\,:\, x_{ij}>0 \}|$.

Second, following \cite{DBLP:conf/coling/JameelS16}, the weighting function $f(x_{ij})$ has been replaced by $\frac{1}{\sigma_j^2}$, where $\sigma_j^2$ is the residual variance of the regression problem for context word $j$, estimated follows:
$$
\sigma_j^2 =  \frac{\sum_{i=1}^n \sum_{j\in J_i}(w_i \cdot \tilde{w_j} + \tilde{b_j} - \textit{PMI}_S(i,j))^2 }{\sum_{i=1}^n |J_i|}
%\sigma_j^2 =  \frac{\sum \{(w_i \cdot \tilde{w_j} + \tilde{b_j} - \textit{PMI}_S(i,j))^2 \,:\, 1\leq i\leq n, j\in J_i \}}{|\{i \,:\, j\in J_i\}|}
$$
Since we need the word vectors to estimate this residual variance, we re-estimate $\sigma_j^2$ after every five iterations of the SGD optimization. For the first 5 iterations, where no estimation for $\sigma_j^2$ is available, we use the GloVe weighting function. 

The use of smoothed frequency counts and residual variance based weighting make the word embedding model more robust for rare words. For instance, if $w$ only co-occurs with a handful of other terms, it is important to prioritize the most informative context words, which is exactly what the use of the residual variance achieves, i.e.\ $\sigma_j^2$ is small for informative terms and large for stop words; see \cite{DBLP:conf/coling/JameelS16}. This will be important for modeling relations, as the relation vectors will often have to be estimated from very sparse co-occurrence counts.

Finally, the bias term $b_i$ has been omitted from the model in \eqref{eqGloVeOurVariant}. We have empirically found that omitting this bias term does not affect the performance of the model, while it allows us to have a more direct connection between the vector $w_i$ and the corresponding PMI scores.

%****************
\subsection{Interpreting Word Vectors in Terms of PMI}\label{secWordVectorsPMI}
Let us define $\textit{PMI}_W$ as follows:
$$
\textit{PMI}_W(i,j) = w_i \cdot \tilde{w_j} + \tilde{b_j}
$$
Then clearly, when the word vectors are trained according to \eqref{eqGloVeOurVariant}, it holds that $\textit{PMI}_W(i,j) \approx \textit{PMI}_S(i,j)$. In other words, we can think of the word vector $w_i$ as a low-rank approximation of the vector $(\textit{PMI}_S(i,1),...,\textit{PMI}_S(i,n))$, with $n$ the number of words in the vocabulary. This view allows us to assign a natural interpretation to some word vector operations. In particular, the vector difference $w_i-w_k$ is commonly used as a model for the relationship between words $i$ and $k$. For a given context word $j$, we have
\begin{align*}
(w_i - w_k) \cdot \tilde{w}_j
&=\textit{PMI}_{W}(i,j)-\textit{PMI}_{W}(k,j)
\end{align*}
The latter is an estimation of
\begin{align*}
\log\left(\frac{P(i,j)}{P(i)P(j)}\right) - \log\left(\frac{P(k,j)}{P(k)P(j)}\right)
=\log \left(\frac{P(j|i)}{P(j|k)}\right) 
\end{align*}
In other words, the vector translation $w_i-w_k$ encodes for each context word $j$ the (log) ratio of the probability of seeing $j$ in the context of $i$ and in the context of $k$, which is in line with the original motivation underlying the GloVe model \cite{glove2014}. In the following section, we will propose a number of alternative vector representations for the relationship between two words, based on generalizations of PMI to three arguments. %These approaches will again crucially rely on the fact that we can directly interpret the word vectors in terms of PMI scores.

%******************************************************************************************
\section{Learning Global Relation Vectors}

In this section, we consider the problem of learning a vector $r_{ik}$ that encodes how the source word $i$ and target word $k$ are related. The main underlying idea is that $r_{ik}$ will capture which context words $j$ are most closely associated with the word pair $(i,k)$. Whereas the GloVe model is based on statistics about (\textit{main word}, \textit{context word}) pairs, here we will need statistics on (\textit{source word}, \textit{context word}, \textit{target word}) triples. First, we discuss how co-occurrence statistics among three words can be expressed using generalizations of PMI to three arguments. Then we explain how this can be used to learn relation vectors in natural way.

\subsection{Co-occurrence Statistics for Triples}\label{secTripleCounts}
Let $\mathcal{P}^l_i \subseteq \{1,...,n_l\}$ again be the set of positions from the $l^{\textit{th}}$ sentence corresponding to word $i$. We define:
\begin{align*}
y_{ijk} = \sum_{l=1}^m\sum_{p\in \mathcal{P}^l_i}\sum_{q\in \mathcal{P}^l_j}\sum_{r\in \mathcal{P}^l_k} \textit{weight}(p,q,r)
%y_{ijk} =& | \{(p,q,r,l) \,:\, w^l_p=i, w^l_q=j, w^l_r=k, \\
%&\quad l\in \{1,...,m\},1 \leq p < q < r \leq n_l, r-p \leq W\} | 
\end{align*}
where $\textit{weight}(p,q,r) = \max(\frac{1}{q-p},\frac{1}{r-q})$ if $p<q<r$ and $r-p \leq W$, and $\textit{weight}(p,q,r)=0$ otherwise. In other words, $y_{ijk}$ reflects the (weighted) number of times word $j$ appears between words $i$ and $k$ in a sentence in which $i$ and $k$ occur sufficiently close to each other. Note that $y_{ijk}$ refers to the number of times the words $i,j,k$ appear in that particular order, whereas in $x_{ij}$ refers to the number of times $j$ appears close to $i$, regardless of whether $i$ or $j$ appears first in the sentence. By taking word order into account in this way, we will be able to model asymmetric relationships.

To model how strongly a context word $j$ is associated with the word pair $(i,k)$, we will consider the following two well-known generalizations of PMI to three arguments \cite{van2011two}:
\begin{align*}
\textit{SI}^1(i,j,k)&=\log\left(\frac{P(i,j)P(i,k)P(j,k)}{P(i)P(j)P(k)P(i,j,k)}\right) \\
\textit{SI}^2(i,j,k)&=\log\left(\frac{P(i,j,k)}{P(i)P(j)P(k)}\right)
\end{align*}
where $P(i,j,k)$ is the probability of seeing the word triple $(i,j,k)$ when randomly choosing a sentence and three (ordered) word positions in that sentence within a window size of $W$. In addition we will also consider two ways in which PMI can be used more directly:
\begin{align*}
\textit{SI}^3(i,j,k)&=  \log\left(\frac{P(i,j,k)}{P(i,k)P(j)}\right)\\
\textit{SI}^4(i,j,k)&=  \log\left(\frac{P(i,k |j)}{P(i|j)P(k|j)}\right)
\end{align*}
Note that $\textit{SI}^3(i,j,k)$ corresponds to the PMI between $(i,k)$ and $j$, whereas $\textit{SI}^4(i,j,k)$ is the PMI between $i$ and $k$ conditioned on the fact that $j$ occurs. The measures $\textit{SI}^3$ and $\textit{SI}^4$ are closely related to $\textit{SI}^1$ and $\textit{SI}^2$ respectively\footnote{Note that probabilities of the form $P(i,j)$ or $P(i)$ here refer to marginal probabilities over ordered triples. In contrast, the PMI scores from the word embedding model are based on probabilities over unordered word pairs, as is common for word embeddings.}. In particular, the following identities are easy to show:
%\begin{align*}
%&\textit{PMI}(i,j) + \textit{PMI}(j,k) - \textit{SI}^1(i,j,k) \\
%&\quad = \log\left(\frac{P(i,j)}{P(i)P(j)}\right) + \log\left(\frac{P(j,k)}{P(j)P(k)}\right) - %\log\left(\frac{P(i,j)P(i,k)P(j,k)}{P(i)P(j)P(k)P(i,j,k)}\right) \\
%&\quad = \log\left(\frac{P(i,j,k)}{P(i,k)P(j)}\right)\\
%&\quad = \textit{SI}^3(i,j,k) 
%\end{align*}
%and
%\begin{align*}
%&\textit{SI}^2(i,j,k) - \textit{PMI}(i,j) - \textit{PMI}(j,k) \\
%&\quad\quad=\log\left(\frac{P(i,j,k)}{P(i)P(j)P(k)}\right)  - \log\left(\frac{P(i,j)}{P(i)P(j)}\right) - %\log\left(\frac{P(j,k)}{P(j)P(k)}\right) \\
%&\quad\quad= \log\left(\frac{P(i,j,k)P(j)}{P(i,j)P(j,k)}\right)\\
%&\quad\quad=  \log\left(\frac{P(i,k|j)}{P(i|j)P(k|j)}\right)\\
%&\quad\quad= \textit{SI}^4(i,j,k)
%\end{align*}
\begin{align*}
\textit{PMI}(i,j) + \textit{PMI}(j,k) - \textit{SI}^1(i,j,k) &= \textit{SI}^3(i,j,k) \\
\textit{SI}^2(i,j,k) - \textit{PMI}(i,j) - \textit{PMI}(j,k) &= \textit{SI}^4(i,j,k)
\end{align*}
Using smoothed versions of the counts $y_{ijk}$, we can use the following probability estimates for $\textit{SI}^1(i,j,k)$--$\textit{SI}^4(i,j,k)$:
\begin{align*}
P(i,j,k) &= \frac{y_{ijk} + \alpha}{y_{***} + n^3 \alpha} &      
P(i,j) &= \frac{y_{ij*} + \alpha}{y_{***} + n^2 \alpha} \\
P(i,k) &= \frac{y_{i*k} + \alpha}{y_{***} + n^2 \alpha} &
P(j,k) &= \frac{y_{*jk} + \alpha}{y_{***} + n^2 \alpha} \\
P(i) &= \frac{y_{i**} + \alpha}{y_{***} + n \alpha} &
P(j) &= \frac{y_{*j*} + \alpha}{y_{***} + n \alpha} \\
P(k) &= \frac{y_{**k} + \alpha}{y_{***} + n \alpha}
\end{align*}
%\begin{align*}
%\textit{SI}^1_S(i,j,k)&=\log\left(\frac{(y_{ij*}+n\alpha)(y_{i*k}+n\alpha)(y_{*jk}+n\alpha)(y_{***}+n^3\alpha)}{(y_{i**} + n^2 \alpha)(y_{*j*} + n^2 \alpha)(y_{**k} + n^2 \alpha)(y_{ijk} + \alpha)}\right)\\
%\textit{SI}^2_S(i,j,k)&=\log\left( \frac{(y_{ijk} + \alpha)(y_{***}+n^3\alpha)^2}{(y_{i**} + n^2 \alpha)(y_{*j*} + n^2 \alpha)(y_{**k} + n^2 \alpha)} \right)\\
%\textit{SI}^3_S(i,j,k)&=  \log\left(\frac{(y_{ijk} + \alpha)(y_{***}+n^3\alpha)}{(y_{i*k} + n \alpha)(y_{*j*} + n^2 \alpha)}\right)\\
%\textit{SI}^4_S(i,j,k)&= \log\left(\frac{(y_{ijk} + \alpha)(y_{*j*} + n^2 \alpha)}{(y_{ij*} + n \alpha)(y_{*jk} + n \alpha)}\right)
%\end{align*}
where $y_{ij*}=\sum_k y_{ijk}$, and similar for the other counts. 

For efficiency reasons, the counts of the form $y_{ij*}$, $y_{i*k}$ and $y_{*jk}$ are pre-computed for all word pairs, which can be done efficiently due to the sparsity of co-occurrence counts (i.e.\ these counts will be 0 for most pairs of words), similarly to how to the counts $x_{ij}$ are computed in GloVe. From these counts, we can also efficiently pre-compute the counts $y_{i**}$, $y_{*j*}$, $y_{**k}$ and $y_{***}$. On the other hand, the counts $y_{ijk}$ cannot be pre-computed, since the total number of triples for which $y_{ijk}\neq 0$ is prohibitively high in a typical corpus. However, using an inverted index, we can efficiently retrieve the sentences that contain the words $i$ and $k$, and since this number of sentences is typically small, we can efficiently obtain the counts $y_{ijk}$ corresponding to a given pair $(i,k)$ whenever they are needed.

%************************************************************************************
\subsection{Relation Vectors}\label{secRelationVectorsFoldingIn}
Our aim is to learn a vector $r_{ik}$ that models the relationship between $i$ and $k$. Computing such a vector for each pair of words (which co-occur at least once) is not feasible, given the number of triples $(i,j,k)$ that would need to be considered. Instead, we first learn a word embedding, by optimizing \eqref{eqGloVeOurVariant}. Then, fixing the context vectors $\tilde{w_j}$ and bias terms $b_j$, we learn a vector representation for a given pair $(i,k)$ of interest by solving the following objective: 
\begin{align}\label{eqFoldingInRelation}
\sum_{j\in J_{i,k}}  (r_{ik} \cdot \tilde{w_j} + \tilde{b_j} - \textit{SI}(i,j,k))^2    
\end{align}
where $\textit{SI}$ refers to one of $\textit{SI}^1_S,\textit{SI}^2_S,\textit{SI}^3_S,\textit{SI}^4_S$. 
Note that \eqref{eqFoldingInRelation} is essentially the counterpart of \eqref{eqGloVeAsPMI}, where we have replaced the role of the PMI measure by SI. In this way, we can exploit the representations of the context words from the word embedding model for learning relation vectors. Note that the factor $\frac{1}{\sigma_j^{2}}$ has been omitted. This is because words $j$ that are normally relatively uninformative (e.g.\ stop words), for which $\sigma_j^{2}$ would be high, can actually be very important for characterizing the relationship between $i$ and $k$. For instance, the phrase ``$X$ such as $Y$'' clearly suggests a hyponomy relationship between $X$ and $Y$, but both `such' and `as' would be associated with a high residual variance $\sigma_j^{2}$. The set $J_{i,k}$ contains every $j$ for which $y_{ijk}>0$ as well as a random sample of $m$ words for which $y_{ijk}=0$, where $m= 2 \cdot |\{j : y_{ijk}>0|$. Note that because $\tilde{w_j}$ is now fixed, \eqref{eqFoldingInRelation} is a linear least squares regression problem, which can be solved exactly and efficiently.

The vector $r_{ik}$ is based on the context words that appear between $i$ and $k$, in sentences that contain both words in that order. In the same way, we can learn a vector $s_{ik}$ based on the context words that appear before $i$ and a vector $t_{ik}$ based on context words that appear after $k$, in sentences where $i$ occurs before $k$. Furthermore, we also learn vectors $r_{ki}$, $s_{ki}$ and $t_{ki}$ from the sentences where $k$ occurs before $i$. As the final representation $R_{ik}$ of the relationship between $i$ and $k$, we concatenate the vectors $r_{ik}, r_{ki},s_{ik}, s_{ki}, t_{ik}, t_{ki}$ as well as the word vectors $w_i$ and $w_k$. We write $R_{ik}^l$ to denote the vector that results from using measure  $\textit{SI}^l$ ($l\in \{1,2,3,4\}$).

%In the experiments, we will also consider the combined vector $R_{ik}^{1234}$ defined as follows:
%\begin{align*}
%R_{ik}^{1234} &=     [r_{ik}^1:r_{ik}^2:r_{ik}^3:r_{ik}^4 :r_{ki}^1:r_{ki}^2:r_{ki}^3:r_{ki}^4 : w_i : w_k]
%\end{align*}
%Finally note that $r_{ik}^l$ is only based on the words that appear between $i$ and $k$ in sentences of the corpus. Similarly, we could consider additional vectors that have been estimated from words appearing before $i$ and words appearing after $k$. \todo{Rephrase this, as we actually are using this; redefine the vectors $R_{ik}^l$ accordingly.}

% need for two directions; possibly consider pre and post vectors as well; add word vectors or difference vector.
% give overview of different configurations that will be used

%****************************************************
\section{Experimental Results}

%\subsection{Methodology}
In our experiments, we have used the Wikipedia dump from November 2nd, 2015, which consists of 1,335,766,618 tokens. 
%\begin{itemize}
%\item The Wikipedia dump from November 2nd, 2015, which consists of 1,335,766,618 tokens. 
%\item The English Gigaword corpus\footnote{https://catalog.ldc.upenn.edu/LDC2011T07}, consisting of 1,094,733,691 tokens.
%\item The UMBC webBase corpus\footnote{http://ebiquity.umbc.edu/resource/html/id/351}, consisting of 2,714,554,484 tokens.
%\item The ClueWeb-2012 Category-B\footnote{http://lemurproject.org/clueweb12/} corpus, which consists of 6,030,992,452 tokens.
%\end{itemize}
We have removed punctuations and HTML/XML tags, and we have lowercased all tokens. Words with fewer than 10 occurrences have been removed from the corpus.
%We used the ReVerb\footnote{http://reverb.cs.washington.edu/} tool to extract sentence from the ClueWeb-2012 (Category B) collection. This tool is specifically designed to preprocess ClueWeb, and only considered tokens that occur less than 100 times in the entire collection, to offset the larger size of this collection. 
To detect sentence boundaries, we have used the Apache sentence segmentation tool\footnote{\url{https://opennlp.apache.org/documentation/1.5.3/manual/opennlp.html#tools.sentdetect}}. In all our experiments, we have set the number of dimensions to 300, which was found to be a good choice in previous work, e.g.\ \cite{glove2014}. We use a context window size $W$ of 10 words. The number of iterations for SGD was set to 50. For our model, we have tuned the smoothing parameter \(\alpha\) based on held-out tuning data, considering values from $\{0.1, 0.01, 0.001, 0.0001, 0.00001, 0.000001\}$. We have noticed that in most of the cases the value of \(\alpha\) was automatically selected as 0.00001. To efficiently compute the triples, we have used the Zettair\footnote{\url{http://www.seg.rmit.edu.au/zettair/}} retrieval engine.
%\todo{Was there one value which seemed to be chosen most of the times? If so, it could be interesting to mention this so there is a default value that could be used in future work}

As our main baselines, we use the following strategies, each of which represents the relationship between a source word $i$ and a target word $k$ as a vector:
\begin{description}
\item[Diff] uses the vector difference $w_k-w_i$.
\item[Conc] uses the concatenation of $w_i$ and $w_k$.
\item[Avg]  averages the vector representations of the words occurring in sentences that contain $i$ and $k$. In particular, let $r^{\textit{avg}}_{ik}$ be obtained by averaging the word vectors of the context words appearing between $i$ and $k$ for each sentence containing $i$ and $k$ (in that order), and then averaging the vectors obtained from each of these sentences. Let $s^{\textit{avg}}_{ik}$ and $t^{\textit{avg}}_{ik}$ be similarly obtained from the words occurring before $i$ and the words occurring after $k$ respectively. The considered relation vector is then defined as the concatenation of $r^{\textit{avg}}_{ik}$, $r^{\textit{avg}}_{ki}$, $s^{\textit{avg}}_{ik}$, $s^{\textit{avg}}_{ki}$, $t^{\textit{avg}}_{ik}$, $t^{\textit{avg}}_{ki}$, $w_i$ and $w_k$. 
\end{description}
The Diff baseline corresponds to the common strategy of modeling relations as vector differences, as e.g.\ in \cite{Vylomova2016}. The vector concatenation model is more general, in the sense that relations which are linearly separable in the Diff representation are always linearly separable in the Conc representation, but the converse does not hold in general. On the other hand, the Conc representation uses twice as many dimensions, which may make it harder to learn a good classifier from few examples. The use of concatenations is popular e.g.\ in the context of hypernym detection \cite{baroni2012entailment}. Finally, the Avg baseline is closest to our model, as this vector is built in the same way as $R_{ik}^1$--$R_{ik}^4$, the only difference being that instead of our relation vectors, the average word vectors are used. This baseline will allow us to directly compare how much we can improve relation vectors by deviating from the common strategy of averaging word vectors. 

%------------------------------------
\subsection{Relation Induction}

%\begin{table*}
%\footnotesize
%\centering
%\caption{Results for the relation induction tasks, using our GloVe variant \eqref{eqGloVeAsPMI}.\label{eqResultsRelationInductionOur}}
%\begin{tabular}{cc  cccc ccccc}
%\toprule
%\multicolumn{11}{c}{Wikipedia}\\
%        &   & Diff & Conc & Avg & Pos &  $R^1_{ik}$ & $R^2_{ik}$ & $R^3_{ik}$ & $R^4_{ik}$ & $R^{1234}_{ik}$ \\
% \midrule
%Google  & Acc    & 90.0 & 89.0 & 89.5 &  89.9     & 90.0 & \textbf{92.3} & 90.9 & 90.4 & 90.2    \\
%        & Pre   & 81.6 & 78.7 & 81.0  &  80.8     & 79.9 & 87.1 & 83.2 & 81.1 & 81.5    \\
%        & Rec    & 82.6 & 83.9 & 82.4  & 83.9      & 86.0 & 84.8 & 84.8 & 85.5 & 85.7    \\
%        & F1   & 82.1 & 81.2 & 81.7 &    82.3   & 82.8 & \textbf{85.9} & 84.0 & 83.3 & 83.6    \\
%\midrule
%DiffVec & Acc    & 29.5 & 28.9 & 29.3 & 29.7      & 29.7 & \textbf{31.3} & 30.4 & 30.1 & 30.5    \\
%        & Pre   & 19.6 & 18.7 & 20.0   & 20.4     & 21.5 & 22.9 & 21.9 & 22.3 & 22.1    \\
%        & Rec    & 23.8 & 22.9 & 23.0  & 23.7      & 24.5 & 25.7 & 25.3 & 22.9 & 24.4    \\
%        & F1   & 21.5 & 20.6 & 21.4 &  21.9     & 22.4 & \textbf{24.2} & 23.5 & 22.6 & 23.2    \\
%\bottomrule
%\end{tabular}
%\end{table*}

\begin{table}
\footnotesize
\centering
\caption{Results for the Google Analogy test set \label{tabResultsGoogle}}
\begin{tabular}{|c | ccc cccc|}
\hline
& Diff & Conc  & Avg &  $R^1_{ik}$ & $R^2_{ik}$ & $R^3_{ik}$ & $R^4_{ik}$ \\
 \hline
Acc    & 90.0 & 89.0  &  89.9     & 90.0 & \textbf{92.3} & 90.9 & 90.4    \\
Pre   & 81.6 & 78.7   &  80.8     & 79.9 & 87.1 & 83.2 & 81.1  \\
Rec    & 82.6 & 83.9   & 83.9      & 86.0 & 84.8 & 84.8 & 85.5    \\
F1   & 82.1 & 81.2 &    82.3   & 82.8 & \textbf{85.9} & 84.0 & 83.3    \\
\hline
\end{tabular}
\end{table}

\begin{table}
\footnotesize
\centering
\caption{Results for the DiffVec test set \label{tabResultsDiffVec}}
\begin{tabular}{|c| ccc cccc|}
\hline
& Diff & Conc  & Avg &  $R^1_{ik}$ & $R^2_{ik}$ & $R^3_{ik}$ & $R^4_{ik}$ \\
 \hline
Acc    & 29.5 & 28.9  & 29.7      & 29.7 & \textbf{31.3} & 30.4 & 30.1   \\
Pre   & 19.6 & 18.7    & 20.4     & 21.5 & 22.9 & 21.9 & 22.3  \\
Rec    & 23.8 & 22.9   & 23.7      & 24.5 & 25.7 & 25.3 & 22.9    \\
F1   & 21.5 & 20.6  &  21.9     & 22.4 & \textbf{24.2} & 23.5 & 22.6   \\
\hline
%& Diff & Conc & Avg & Pos &  $R^1_{ik}$ & $R^2_{ik}$ & $R^3_{ik}$ & $R^4_{ik}$ \\
% \midrule
%Acc    & 90.0 & 89.0 & 89.5 &  89.9     & 90.0 & \textbf{92.3} & 90.9 & 90.4    \\
%Pre   & 81.6 & 78.7 & 81.0  &  80.8     & 79.9 & 87.1 & 83.2 & 81.1  \\
%Rec    & 82.6 & 83.9 & 82.4  & 83.9      & 86.0 & 84.8 & 84.8 & 85.5    \\
%F1   & 82.1 & 81.2 & 81.7 &    82.3   & 82.8 & \textbf{85.9} & 84.0 & 83.3    \\
%\midrule
%Acc    & 29.5 & 28.9 & 29.3 & 29.7      & 29.7 & \textbf{31.3} & 30.4 & 30.1   \\
%Pre   & 19.6 & 18.7 & 20.0   & 20.4     & 21.5 & 22.9 & 21.9 & 22.3  \\
%Rec    & 23.8 & 22.9 & 23.0  & 23.7      & 24.5 & 25.7 & 25.3 & 22.9    \\
%F1   & 21.5 & 20.6 & 21.4 &  21.9     & 22.4 & \textbf{24.2} & 23.5 & 22.6   \\
%\bottomrule
\end{tabular}
\end{table}

In the relation induction task, we are given a set of word pairs $(s_1,t_1),...,(s_k,t_k)$ that are related in some way, and the task is to decide for a number of test examples $(s,t)$ whether they also have this relationship. Among others, this task was considered in \cite{Vylomova2016}, and a ranking version of this task was studied in \cite{DBLP:conf/coling/DrozdGM16}. 

We will use two different test sets. First, we consider the Google Analogy Test Set \cite{DBLP:journals/corr/abs-1301-3781}. This dataset contains instances of 14 different types of relations. Second, we use the DiffVec dataset, which was introduced in \cite{Vylomova2016}. This dataset contains instances of 36 different types of relations. Note that both datasets contain a mix of semantic and syntactic relations. 
%Finally, we consider a test set which we derived from the webtable dataset\footnote{\url{http://webdatacommons.org/webtables/}} \cite{lehmberg2016large}. From this dataset, we have only considered the relational tables, as the rows of these tables often express some relationship between different entities. To obtain interesting tables, i.e.\ tables that are likely to express a meaningful relationship, we applied three filters. First, since we are focusing on binary relationships, we have restricted ourselves to tables that either contain two columns, or contain three columns but in which the first column is simply an enumeration. Second, we have selected those tables in which at least 50\% of all entries in either column map to WikiData\footnote{\url{https://www.wikidata.org}} entities of the same semantic type. The underlying motivation is that tables which express a meaningful relationship will typically indeed have columns that are semantically homogeneous. Finally, from the resulting set of tables, we have selected those that have at least 50 rows that consist of words from the word embedding vocabulary. This resulted in a total of 3056 tables.

In our evaluation, we have used 10-fold cross-validation (or leave-on-out for relations with fewer than 10 instances). In the experiments, we consider for each relation in the test set a separate binary classification task, which was found to be considerably more challenging than a multi-class classification setting in \cite{Vylomova2016}. To generate negative examples in the training data (resp.\ test data), we have used three strategies, following \cite{Vylomova2016}. First, for a given positive example $(s,t)$ of the considered relation, we add $(t,s)$ as a negative example. Second, for each positive example $(s,t)$, we generate two negative examples $(s,t_1)$ and $(s,t_2)$ by randomly selecting two tail words $t_1, t_2$ from the other training (resp.\ test) examples of the same relation. Finally, for each positive example, we also generate a negative example by randomly selecting two words from the vocabulary. For each relation, we then train a linear SVM classifier. To set the parameters of the SVM, we initially use 25\% of the training data for tuning, and then retrain the SVM with the optimal parameters on the full training data.

The results are summarized in Table \ref{tabResultsGoogle} for the Google analogy dataset and in Table \ref{tabResultsDiffVec} for DiffVec, in terms of accuracy and (macro-averaged) precision, recall and F1 score. As can be observed, our model outperforms the baselines on both datasets for both accuracy and F1 score, with the $R_{ik}^2$ variant outperforming the others. Regarding the baselines, it is particularly noteworthy that the Avg baseline is barely able to outperform Diff. This confirms the findings from earlier work that the relations in the Google Analogy and DiffVec datasets can be modelled rather well using a vector translation model \cite{Vylomova2016}. % \todo{We need some examples here of relations where we do much better than the baselines + some discussion on whether the improvement is consistent or whether there are also relations where the baselines are better. I would guess that for relations with few examples, we will sometimes do worse due to the high dimensionality of our relation vectors.}

\begin{table}
\footnotesize
\centering
\caption{Results without position weighting. \label{tabWeighting}}
\begin{tabular}{|c | cc|cc|}
\hline
& \multicolumn{2}{c}{Google}  & \multicolumn{2}{c|}{DiffVec} \\
\hline
& Acc &  F1 & Acc & F1  \\ \hline
$R^1_{ik}$ & 89.7 & 82.4 & 30.2 & 22.2  \\
$R^2_{ik}$ & 91.0 & 83.4 & 30.8 & 24.1 \\
$R^3_{ik}$ & 90.4 & 83.2 & 30.1 & 22.3 \\
$R^4_{ik}$ & 90.2 & 82.9 & 29.1 & 21.2 \\
\hline
\end{tabular}
\end{table}

\begin{table}
\footnotesize
\centering
\caption{Results without the relation vectors $s_{ik}$ and $t_{ik}$.\label{tabBeforeAfter}}
\begin{tabular}{|c | cc|cc|}
\hline
& \multicolumn{2}{c}{Google}  & \multicolumn{2}{c|}{DiffVec} \\
\hline
& Acc &  F1 & Acc & F1  \\ \hline
$R^1_{ik}$ & 90.0 & 82.5 & 29.9  & 22.3  \\
$R^2_{ik}$ & 92.3 & 85.8 & 31.2 & 24.2 \\
$R^3_{ik}$ & 90.5 & 83.2 & 30.2 & 23.0 \\
$R^4_{ik}$ & 90.3 & 83.1 & 29.8 & 22.3 \\
\hline
\end{tabular}
\end{table}

Similar as in the GloVe model, the context words in our model are weighted based on their distance to the nearest target word. Table \ref{tabWeighting} shows the results for our model without this weighting. Comparing these results with those in Tables \ref{tabResultsGoogle} and \ref{tabResultsDiffVec} shows that the weighting scheme indeed leads to a small improvement (except for the accuracy of $R_{ik}^1$ for DiffVec). Similarly, in Table \ref{tabBeforeAfter}, we show what happens if the relation vectors $s_{ik}$, $s_{ki}$, $t_{ik}$ and $t_{ki}$ are omitted. In other words, for the results in Table \ref{tabBeforeAfter}, we only use context words that appear between the two target words. Again, in general the results are worse than those in Tables \ref{tabResultsGoogle} and \ref{tabResultsDiffVec} (with the accuracy of $R_{ik}^1$ for DiffVec again being an exception), although the differences are very small in this case. While including the vectors $s_{ik}$, $s_{ki}$, $t_{ik}$, $t_{ki}$ should be helpful, it also significantly increases the dimensionality of the vectors $R_{ik}^l$. Given that the number of instances per relation is typically quite small for this task, this can also make it harder to learn a suitable classifier.

\begin{table*}
\footnotesize
\centering
\caption{Results for the relation induction task using alternative word embedding models.\label{tabResultsAlternativeWordEmbedding}}
\begin{tabular}{|c | cc cc | cc cc| cc cc|}
\hline
& \multicolumn{4}{c|}{GloVe} & \multicolumn{4}{c|}{SkipGram} & \multicolumn{4}{c|}{CBOW}\\
\hline
& \multicolumn{2}{c}{Google}  & \multicolumn{2}{c|}{DiffVec} & \multicolumn{2}{c}{Google}  & \multicolumn{2}{c|}{DiffVec} & \multicolumn{2}{c}{Google}  & \multicolumn{2}{c|}{DiffVec}\\
\hline
& Acc &  F1  & Acc & F1 & Acc &  F1 & Acc &  F1 & Acc &  F1 & Acc  & F1\\
Diff & 90.0  & 81.9 & 21.2  & 13.9 & 89.8  & 81.9 & 21.7  & 14.5 & 89.9 & 82.1 & 17.4 & 9.7 \\
Conc & 88.9&  80.4 & 20.2& 11.9 & 89.2& 81.6 & 20.5& 12.0 & 89.1 & 81.1 & 16.4 & 7.7\\
Avg & 89.8 & 82.1 & 21.4& 13.9 & 90.2 & 82.4 & 21.8& 14.4 & 89.8 & 82.2 & 17.5 & 10.0\\
$R^1_{ik}$ & 89.7 &  81.7 & 20.9 & 12.5 & 89.4 &  81.2 & 21.1 & 12.3 & 89.8 & 81.9 & 17.2 & 9.2\\
$R^2_{ik}$ & 90.0 &  82.8 & 21.2 & 13.4 & 89.1 &  81.3 & 21.1 &  12.9 & 90.2 & 82.4 & 17.7 & 10.0\\
$R^3_{ik}$ & 90.0 &  82.3 & 20.0 & 11.2 & 89.5&  81.1 & 20.5 & 12.3 & 89.5 & 81.1 & 17.2 & 9.6 \\
$R^4_{ik}$ & 90.0 &  82.5 & 20.0 & 11.4 & 88.9 &  80.8 & 20.6 & 12.1 & 90.5 & 82.2 & 17.1 & 8.4 \\
\hline
\end{tabular}
\end{table*}

Finally, to analyze the benefit of our proposed word embedding variant, Table \ref{tabResultsAlternativeWordEmbedding} shows the results that were obtained when we use a standard word embedding model. In particular, we show results for the standard GloVe model, SkipGram and the Continuous Bag of Words (CBOW) model. As can be observed, our variant leads to better results than the original GloVe model, both for the baselines and for our relation vectors. The difference is particularly noticeable for DiffVec. The difference is also larger for our relation vectors than for the baselines, which is expected as our method is based on the assumption that context word vectors can be interpreted in terms of PMI scores, which is only true for our variant. For SkipGram, we see a similar drop in performance on the DiffVec test set, but not for the baselines on the Google Analogy test set, which is perhaps not sufficiently challenging to see clear differences. Finally, the results for CBOW are similar to these for GloVe and SkipGram for the Google Analogy test set, but they are worse in the case of DiffVec. These results show that the proposed changes to the GloVe model indeed lead to a meaningful improvement. In the remainder of this paper, we will therefore only consider our variant.

%------------------------------------
\subsection{Measuring Degrees of Prototypicality}
Instances of relations can often have different degrees of prototypicality. For example, for the relation ``$X$ characteristically makes the sound $Y$'', the pair (\textit{dog},\textit{bark}) should be considered more prototypical than the pair (\textit{floor},\textit{squeak}), even though both pairs might be considered to be instances of the relation in a classification setting \cite{jurgens2012semeval}. A suitable relation vector should allow us to rank word pairs according to how prototypical they are as instances of that relation. To evaluate the ability of our relation vectors to produce such rankings, we use a dataset that was produced in the aftermath of SemEval 2012 Task 2, where a closely related problem was considered. In particular, we have used the ``Phase2AnswerScaled'' data from the platinum rankings dataset, which is available from the SemEval 2012 Task 2 website\footnote{https://sites.google.com/site/semeval2012task2/download}. In this dataset, 79 ranked list of word pairs are provided, each of which corresponds to a particular relation. We then considered the following experimental setting. For each relation, we first split the associated ranking into 60\% training, 20\% tuning, and 20\% testing (i.e.\ we randomly select 60\% of the word pairs and use their ranking as training data, and similar for tuning and test data). We then train a linear SVM regression model (using SVMLight\footnote{http://svmlight.joachims.org/}) on the ranked word pairs from the training set and use the tuning set to tune the parameters of the model. 

\begin{table}
\footnotesize
\centering
\caption{Results for measuring degrees of prototypicality (Spearman $\rho \times 100$). \label{eqResultsDiffVec}}
\begin{tabular}{| c c c  c c c c |}
\hline
  Diff & Conc  & Avg & $R^1_{ik}$ & $R^2_{ik}$ & $R^3_{ik}$ & $R^4_{ik}$\\ 
 \hline
  17.3 & 16.7 & 21.1 & 22.7 & \textbf{23.9} & 21.8 & 22.2\\
 \hline
\end{tabular}
\end{table}

\begin{figure}
\centering
\includegraphics[scale=1]{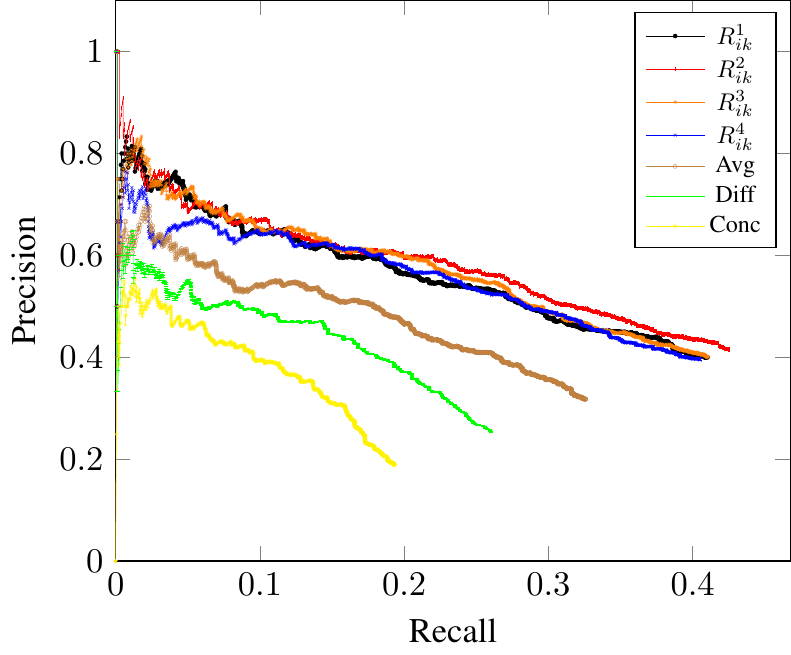}
\caption{Results for the relation extraction from the NYT corpus: comparison with the main baselines. \label{figNYTourBaselines}}
\end{figure}

We report result using Spearman's $\rho$ in Table \ref{eqResultsDiffVec}. We can observe that our model again outperforms the baselines, with $R_{ik}^2$ again being the best performing variant. Interestingly, in this case, the Avg baseline is considerably stronger than Diff and Conc. Intuitively, we might indeed expect that this ranking problem requires a more fine-grained representation than the classification setting from the previous experiment.

%------------------------------------
\subsection{Relation Extraction}
Finally, we consider the problem of relation extraction from a text corpus. Specifically, we consider the task proposed in \cite{DBLP:conf/pkdd/RiedelYM10}, which is to extract (subject,predicate,object) triples from the New York Times (NYT) corpus. Rather than having labelled sentences as training data, we have to use the existing triples from Freebase as a form of distant supervision, i.e.\ for some pairs of entities we know some of the relations that hold between them, but not which sentences assert these relationships (if any). To be consistent with published results for this task, we have used a word embedding that was trained from the NYT corpus\footnote{https://catalog.ldc.upenn.edu/LDC2008T19}, rather than Wikipedia (using the same preprocessing and set-up). We have used the training and test data that was shared publicly for this task\footnote{http://iesl.cs.umass.edu/riedel/ecml/}, which consist of sentences from articles published in 2005-2006 and in 2007, respectively. Each of these sentences contains two entities, which are already linked to Freebase. 
We learn relation vectors from the sentences in the training and test sets, and learn a linear SVM classifier based on the Freebase triples that are available in the training set. Initially, we split the training data into 75\% training and 25\% tuning to find the optimal parameters of the linear SVM model. After tuning, we re-train the SVM models on the full training data. 

\begin{figure}
\centering
\includegraphics[scale=1]{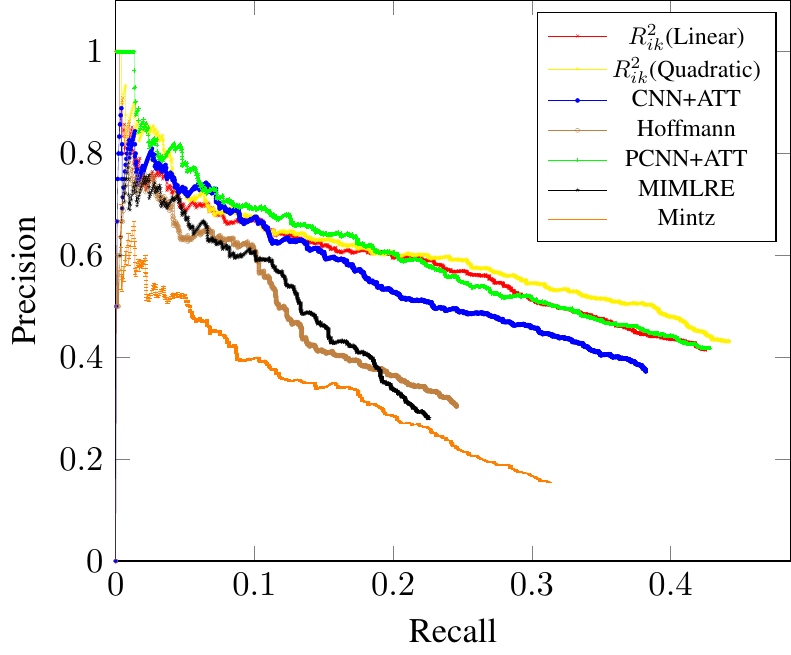}
\caption{Results for the relation extraction from the NYT corpus: comparison with state-of-the-art neural network models. \label{figNYTNN}}
\end{figure}

Following earlier work on this task, we report our results on the test set as a precision-recall graph in Figure \ref{figNYTourBaselines}. This shows that the best performance is again achieved by $R_{ik}^2$, especially for larger recall values. Note that the differences between the baselines are more pronounced in this task, with Avg being clearly better than Diff, which is in turn better than Conc. For this relation extraction task, a large number of methods have already been proposed in the literature, with variants of convolutional neural network models with attention mechanisms achieving state-of-the-art performance. A comparison with these models\footnote{Results for the neural network models have been obtained from \url{https://github.com/thunlp/TensorFlow-NRE/tree/master/data}.} is shown in Figure \ref{figNYTNN}. As can be observed from Figure \ref{figNYTNN}, the performance of $R_{ik}^2$ is comparable with the PCNN+ATT model  \cite{DBLP:conf/acl/LinSLLS16}, especially for larger recall values, and it outperforms the remaining models. This is remarkable, as our model is conceptually much simpler, and has not been designed specifically for this task. For instance, a version of the attention mechanism that is used in the PCNN+ATT model could easily be incorporated into our model, which might lead to further improvements. Finally, we also tested our model with an SVM classifier with a quadratic kernel, as the amount of training data is larger than for the previous tasks. As can be seen in Figure \ref{figNYTNN}, this leads to a clear improvement, with better results than PCNN+ATT for larger recall values.

%****************************************************
\section{Conclusions}
We have proposed an unsupervised method which uses global co-occurrences statistics from a text corpus to represent the relationship between a given pair of words as a vector. To this end, we have first proposed a variant of the GloVe word embedding model, in which word vectors explicitly correspond to smoothed PMI-weighted co-occurrence vectors. Our relation vectors are then obtained in a similar way, by using generalizations of PMI to three arguments. In contrast to neural network models for relation extraction, our model learns relation vectors in an unsupervised way. Moreover, even in (distantly) supervised tasks (where we need to learn a classifier on top of the unsupervised relation vectors), our model has proven competitive with state-of-the-art neural network models. Compared to approaches that rely on averaging word vectors, our method is able to learn more faithful representations by focusing on the words that are most strongly related to the considered relationship. Compared to vector difference based methods, the advantage of our method is that it can also model relationships that are not determined by the salient features of the words themselves. 

\newpage
\bibliography{commonsense,wordembedding}
\bibliographystyle{aaai}

\end{document}